\documentclass[conference]{IEEEtran}
\IEEEoverridecommandlockouts
\usepackage{cite}
\usepackage{amsmath,amssymb,amsfonts}
\usepackage{algorithmic}
\usepackage{graphicx}
\usepackage{textcomp}
\usepackage{xcolor}
\usepackage{booktabs}
\usepackage{graphicx}
\usepackage{subcaption}
\usepackage{hyperref}
\usepackage{svg}

\newcommand{\method}{\textit{Pre-VLA}} %
\def\BibTeX{{\rm B\kern-.05em{\sc i\kern-.025em b}\kern-.08em
    T\kern-.1667em\lower.7ex\hbox{E}\kern-.125emX}}
\begin{document}
     
\title{\method{}: Preemptive Runtime Verification for Reliable Vision-Language-Action and World-Model Rollouts
\\ % 随便先取个名字占位
% {\footnotesize \textsuperscript{*}Note: Sub-titles are not captured for https://ieeexplore.ieee.org  and
% should not be used
% }
% \thanks{Identify applicable funding agency here. If none, delete this.}
}

\author{
\IEEEauthorblockN{Zhen Sun$^{1,4}$\textsuperscript{*}, Yongjian Guo$^{2,4}$\textsuperscript{*}, Haoran Sun$^{3}$, Luqiao Wang$^{1,4}$, Wei Lu$^4$, \\
Jiachi Ji$^1$, Shengzhe Ji$^5$, Junwu Xiong$^4$\textsuperscript{\dag}, Zhijun Meng$^1$\textsuperscript{\dag}}
\IEEEauthorblockA{$^1$Beihang University, $^2$Tsinghua University, $^3$Peking University, \\
$^4$JDT AI Infra, $^5$Zhejiang University}

\thanks{\textsuperscript{*}Equal Contribution}
%\thanks{\textsuperscript{\dag} Corresponding Author: Zhijun Meng, Junwu Xiong }
\thanks{\textsuperscript{\dag} Corresponding Author: Junwu Xiong, Zhijun Meng }

}
% \author{\IEEEauthorblockN{1\textsuperscript{st} Given Name Surname}
% \IEEEauthorblockA{\textit{dept. name of organization (of Aff.)} \\
% \textit{name of organization (of Aff.)}\\
% City, Country \\
% email address or ORCID}
% \and
% \IEEEauthorblockN{2\textsuperscript{nd} Given Name Surname}
% \IEEEauthorblockA{\textit{dept. name of organization (of Aff.)} \\
% \textit{name of organization (of Aff.)}\\
% City, Country \\
% email address or ORCID}
% \and
% \IEEEauthorblockN{3\textsuperscript{rd} Given Name Surname}
% \IEEEauthorblockA{\textit{dept. name of organization (of Aff.)} \\
% \textit{name of organization (of Aff.)}\\
% City, Country \\
% email address or ORCID}
% \and
% \IEEEauthorblockN{4\textsuperscript{th} Given Name Surname}
% \IEEEauthorblockA{\textit{dept. name of organization (of Aff.)} \\
% \textit{name of organization (of Aff.)}\\
% City, Country \\
% email address or ORCID}
% \and
% \IEEEauthorblockN{5\textsuperscript{th} Given Name Surname}
% \IEEEauthorblockA{\textit{dept. name of organization (of Aff.)} \\
% \textit{name of organization (of Aff.)}\\
% City, Country \\
% email address or ORCID}
% \and
% \IEEEauthorblockN{6\textsuperscript{th} Given Name Surname}
% \IEEEauthorblockA{\textit{dept. name of organization (of Aff.)} \\
% \textit{name of organization (of Aff.)}\\
% City, Country \\
% email address or ORCID}
% }

\maketitle

\begin{abstract}
While large vision-language-action (VLA) models and generative world models (WM) have advanced long-horizon embodied intelligence, their practical deployment remains challenged by uncertainty in learning-based action generation. Low-quality actions may cause physical failures during execution or lead to misleading world-model rollouts with redundant rendering costs. To address this issue, we propose \method{}, a unified runtime verification architecture that performs preemptive action validity assessment before physical execution or world-model imagination. \method{} leverages an efficient multimodal backbone with modality-aware pooling and a lightweight dual-branch head to predict both safety confidence and critic-derived advantage scores for candidate action chunks. To handle severe class imbalance and unstable boundary decisions, we train \method{} with a multi-task objective combining Focal classification, advantage regression, and soft-threshold calibration. During deployment, a dual-mode preemptive resampling scheduler filters low-quality actions and triggers adaptive resampling under a limited computation budget. Experiments on the LIBERO benchmark show that \method{} improves the average closed-loop success rate across four suites from 30.79\% to 37.62\% over RynnVLA-002, reduces task execution steps, achieves 183.9 ms average forward verification time per action chunk, and mitigates error accumulation in world-model rollouts.

\end{abstract}


\section{Introduction}
\label{sec:intro}

% In embodied intelligent long-horizon operational tasks, large vision-language-action (VLA) models~\cite{kim2025openvla,bjorck2025gr00t} and action-conditioned world models (WM) are widely employed to generate control policies and predict physical futures. However, in resource-constrained robotic control systems, directly executing subpar actions from VLA can lead to irreversible physical failures~\cite{todo}, while feeding these actions into WM for forward simulation may trigger severe physical hallucinations~\cite{gao2026sword} and waste highly expensive computational budgets~\cite{todo}. To address this, we propose a unified runtime verification architecture-\method{}. \method{} utilizes lightweight multimodal monitors to preemptively intercept physical executions or state predictions, thereby ensuring both the physical safety of the system and the efficient utilization of computational resources. 

In embodied intelligent long-horizon operational tasks, large vision-language-action (VLA) models~\cite{kim2025openvla,bjorck2025gr00t} and action-conditioned world models (WM)~\cite{wan2025wan,huang2026noisegate} have been widely employed to generate control policies and predict complex physical futures~\cite{ma2024survey,zhong2025survey,zhang2025pure}. By transforming heterogeneous multimodal sensory streams into unified discrete or continuous representations, these foundational models endow robotic agents with remarkable semantic generalization capabilities and abstract reasoning skills across various open-world environments~\cite{zhou2026thousand}. In parallel, generative world models serve as neural simulators that synthesize predictive visual futures conditioned on candidate actions, thereby enabling proactive planning and multi-step lookahead verification~\cite{jiang2026wovr}. Despite these promising advances, existing embodied models still face the challenge of learning-based generative uncertainty during practical deployment~\cite{feng2025multiagentembodiedaiadvances}. At each planning cycle, the candidate actions generated by the VLA policy may be affected by distributional shifts, temporal error accumulation, or insufficient confidence calibration~\cite{cen2025worldvla}. Once low-quality actions enter the downstream pipeline, they may introduce risks in two scenarios: in physical execution, they can cause collisions, object drops, or irreversible kinematic violations; in world model imagination, although the WM can produce relatively stable future predictions under reasonable action conditions, biased or unsafe action inputs may lead to blurry frames, target position drift, distorted physical relationships, or even false-success rollouts~\cite{jiang2026world4rldiffusionworldmodels}, causing visual and physical errors to accumulate while incurring additional GPU rendering costs~\cite{gao2026sword,guan2026rl}. Therefore, establishing a runtime verification mechanism capable of preemptive safety assessment and candidate action filtering has become a fundamental prerequisite for trustworthy embodied intelligence.

\begin{figure}
    \centering
    \includegraphics[width=1\linewidth]{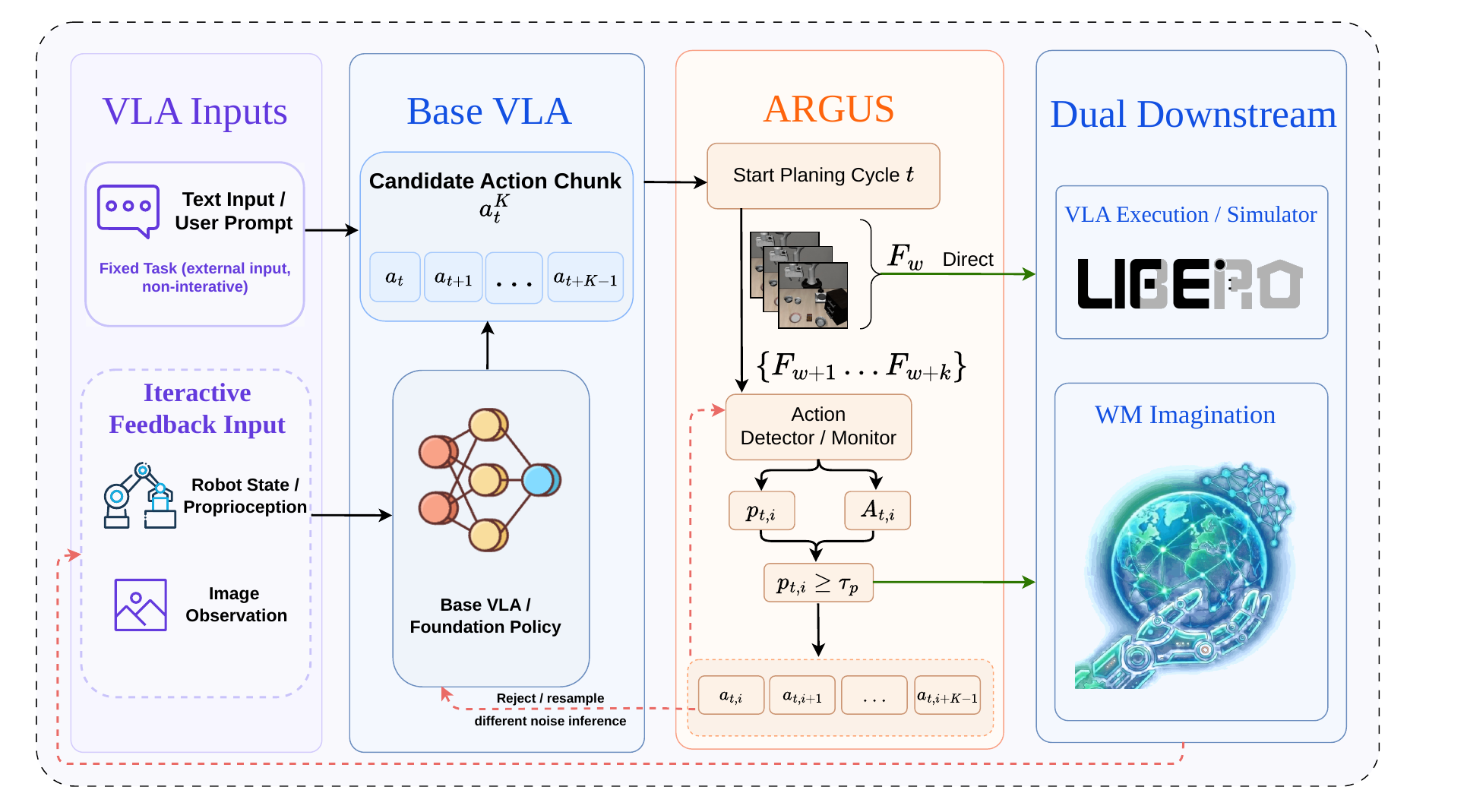}
    \caption{Overview of the dual-mode runtime verification pipeline of \method{}.}
    \label{fig:framework}
\end{figure}

However, designing an effective runtime verification framework for highly dynamic robotic systems cannot be achieved by simply adding a filtering module, as it faces several challenges~\cite{caldas2024rosrv}. First, robotic tasks typically involve multimodal information, including language, vision, proprioceptive states, and actions. Improper handling of such high-dimensional heterogeneous data can introduce considerable computational overhead~\cite{guan2024search}; for robotic control pipelines that require continuous operation, excessive verification latency may affect the stability of action execution~\cite{betzer2024digitaltwin}. Second, existing embodied runtime evaluation modules mainly focus on evaluating environmental states or prediction results, making it difficult to directly assess the safety and validity of candidate actions before they enter physical execution or world model imagination~\cite{wang2025robosafe}. Therefore, constructing a lightweight pre-verification mechanism for action chunks is crucial for reliable runtime safety assurance.In addition, real-world offline exploration data is often limited~\cite{bjorck2025gr00t} and suffers from severe class imbalance. Since standard safe trajectories dominate the data distribution~\cite{xu2025wod,jiang2026wovr}, rare yet critical catastrophic failure samples can be overwhelmed by abundant safe samples during standard cross-entropy optimization~\cite{visca2021deep,zhao2025survey}. This imbalance weakens the model's ability to identify high-risk actions and makes it more prone to over-confidence or decision oscillations near safety boundaries.

To address the above limitations, we propose \method{}
, a unified runtime verification architecture that seamlessly integrates runtime safety constraints into the foundational inference pipeline of embodied agents, as shown in Fig~\ref{fig:framework}. The core idea of \method{} is to leverage the efficient unified multimodal backbone of the advanced WorldVLA~\cite{cen2025worldvla} network to jointly represent language instructions, visual observations, proprioceptive states, and candidate actions. By introducing dedicated action and state tokenizers, \method{} encodes multi-source inputs within a unified text-image representation space and extracts the final-layer hidden states of the backbone as a low-overhead feature basis for subsequent action validity verification.To handle the complex coupling among heterogeneous features, we design an explicit modality-aware pooling layer, which uses predefined masks to separate the independent contributions of text instructions, visual observations, proprioceptive states, and action chunks. The concatenated multimodal feature vector is then fed into a minimalist dual-branch feed-forward network. The primary classification branch outputs a safety confidence score, while the parallel regression branch predicts a continuous long-horizon advantage score derived from the Critic. To address the prevalent dataset imbalance and mitigate boundary oscillations, we construct a multi-task joint optimization objective that combines Focal Loss~\cite{lin2017focal} for hard-example-focused classification, a mean squared error regression loss for physical regularization, and an adaptive Soft-threshold Loss for smooth probability calibration. During physical execution or world model imagination, the dual-modal preemptive resampling scheduler leverages the above dual-branch predictions to dynamically intercept anomalous action chunks before they trigger unrecoverable kinematic violations or expensive simulation rendering loops. In this way, \method{} effectively maintains the system safety boundary under a limited resampling budget.

We conduct a comprehensive evaluation of \method{} on the LIBERO~\cite{liu2023libero} robotic manipulation benchmark, covering action validity discrimination, closed-loop physical execution, world model rollouts, ablation studies, and task-level case studies under stronger baseline settings. Experimental results show that \method{} achieves an F1 score of 0.8303 and an accuracy of 0.9542 on the independent LIBERO test set, while reducing the false pass rate of invalid actions to 0.0200, indicating its ability to distinguish valid actions from high-risk candidates. In closed-loop execution experiments across four LIBERO suites, \method{} improves the average success rate from 30.79\% to 37.62\% compared with RynnVLA-002~\cite{cen2025rynnvla}, yielding an absolute gain of 6.83 percentage points. Meanwhile, it reduces the number of task execution steps, suggesting that action filtering helps the policy enter valid execution trajectories more efficiently. Runtime analysis further shows that the average forward verification time of \method{} for a single candidate action chunk is approximately 183.9 ms, indicating manageable computational overhead. In addition, qualitative world model rollout results on LIBERO tasks demonstrate that \method{} can filter low-quality actions before they enter the WM, thereby mitigating misleading future predictions, target drift, and visual artifacts. Finally, ablation studies and high-baseline task-level case studies on LIBERO further validate the effectiveness of the multi-task training objective and probability threshold calibration, showing that \method{} can provide stable gains in both task success and execution efficiency even when applied to stronger base policies.
The main contributions of this work are presented as follows. 
\begin{itemize}
    \item We introduce \method{}, a unified runtime verification architecture for embodied intelligence that preemptively secures physical robotic actuators and world model simulators via an ultra-low-latency filtering pipeline. 
    
    \item We design a multimodal action verification module built upon an efficient unified backbone. By incorporating modality-aware pooling and a dual-branch prediction head, the module jointly encodes language instructions, visual observations, proprioceptive states, and candidate action chunks, while simultaneously predicting action safety confidence and action advantage scores. In addition, we formulate a multi-task training objective that combines Focal classification, advantage regression, and soft-threshold calibration to mitigate class imbalance and improve decision stability near safety boundaries. 
    
    \item Through rigorous evaluation on the LIBERO~\cite{liu2023libero} manipulation benchmark, we demonstrate that \method{} achieves remarkable improvements in VLA closed-loop execution success rates, substantial reductions in environment interaction steps, and robust mitigation of world model hallucinations under strict runtime computational budgets.
\end{itemize}

\section{Related Work}
\textbf{Vision-Language-Action Models and Generative World Models.} 
Recent advancements in embodied intelligence have largely benefited from the synergistic evolution of large-scale Vision-Language-Action (VLA) models and action-conditioned World Models. Foundational models such as OpenVLA~\cite{kim2025openvla}, $\pi$~\cite{black2024pi_0, intelligence2025pi_}, and GR00T~\cite{bjorck2025gr00t} have demonstrated remarkable semantic understanding and task generalization capabilities through pre-training on internet-scale multimodal datasets. To overcome the prohibitive costs of real-world data collection, researchers have explored utilizing world models as neural simulators~\cite{jiang2026wovr} to drive policy optimization via "imagination." CTRL-World~\cite{guo2025ctrl} introduced a controllable multi-view generation mechanism, providing high-fidelity environmental feedback for evaluating the instruction-following capabilities of generalist policies. Subsequently, GigaBrain-0.5M~\cite{team2026gigabrain} further validated the potential of world-model-based reinforcement learning in handling complex manipulation tasks. WMPO~\cite{WMPO2025} achieved on-policy optimization without real-world environment interaction by aligning predicted pixels with pre-trained VLA features. However, the fidelity of generative simulators remains constrained by physical consistency challenges. World-VLA-Loop~\cite{liu2026world} attempted to mitigate this through closed-loop joint optimization of world models and policies, while Sword~\cite{gao2026sword} significantly enhanced the robustness of world models under distribution shifts via structure-guided style augmentation. %Nevertheless, as noted in WoVR~\cite{jiang2026wovr}, inevitable cascading hallucinations and long-horizon error accumulation in closed-loop imagination remain core bottlenecks restricting the reliability of such systems. This has prompted the research community to re-examine how to identify and rectify potential anomalous actions in real-time during the inference phase.

\textbf{Runtime Verification in Embodied AI.} 
As embodied intelligent systems enter dynamic and unpredictable real-world scenarios, ensuring that their behaviors adhere to physical constraints and safety boundaries has become a focal point of academic attention. Traditional robotic safety research primarily focuses on formal verification or control-theory-based defense barriers~\cite{ames2019control, luo2021learning}, which face the curse of dimensionality when processing high-dimensional sensory inputs. Within learning-driven paradigms, early attempts mainly involved adding discriminative filters to identify unsafe states~\cite{dalal2018safe,deng2025ai}, but these were often limited by computational latency and perceptual redundancy~\cite{bharadhwaj2020conservative}. 
%Most of these methods serve as post-hoc remedies and lack deep perception of the underlying physical dynamics. Furthermore, existing runtime monitoring schemes perform poorly when faced with the extreme data imbalance inherent in the real world, particularly where safe samples overwhelmingly dominate fatal failure samples; in such cases, standard cross-entropy optimization easily leads to a high miss rate for high-risk actions. 
To address generative uncertainty in diffusion or flow-matching models, NoiseGate~\cite{huang2026noisegate} proposed a timestep scheduling strategy based on information gating, optimizing the reliability of action generation by regulating the latent noise distribution~\cite{chen2024diffusion}.
Additionally, conventional hard-thresholding discriminative models ignore the continuous evolutionary characteristics of action quality, failing to effectively utilize the advantage information provided by reinforcement learning algorithms such as Proximal Policy Optimization PPO~\cite{schulman2017proximal}.

% \section{Methodology}
% \subsection{overview of \method{}}
% \subsection{\method{}: process1}
% \subsection{\method{}: process2}
% \subsection{\method{}: process3}

\begin{figure*}[h!]
    \centering
    \includegraphics[width=1\linewidth]{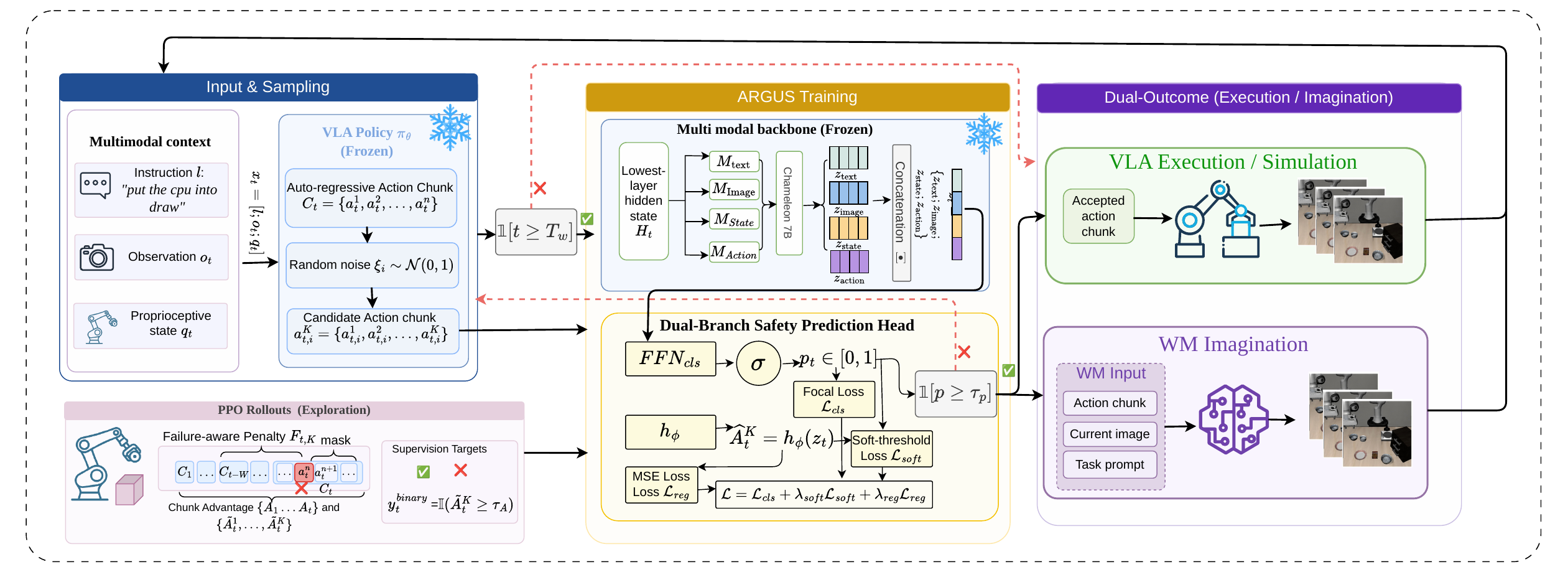}
    \caption{Overview of the ARGUS runtime safety verification framework. The VLA generates candidate action chunks from instructions, observations, and proprioceptive states. ARGUS learns a safety verifier from PPO rollouts to filter, resample, and select fallback actions at runtime before dispatching verified actions to physical execution/simulation or world-model imagination.}
    \label{fig:pipeline}
\end{figure*}

\section{Runtime Safety Verification Framework: \method{}}

\subsection{Overview of \method{}}
\method{} supports dual-mode deployment, consisting of a base VLA policy and an action-conditioned world model, as shown in Fig.~\ref{fig:pipeline}. At each planning cycle $t$, the VLA autoregressively generates a candidate action chunk $a_t^K$ from the multimodal context. Before the action chunk is either executed on the physical robot or used by the WM for next-frame prediction, it is first evaluated by \method{}.

\begin{itemize}
    \item \textbf{VLA Execution in physical actuator:} Intercepts dangerous actions to prevent physical collisions. Since the base model lacks out-of-distribution (OOD) recovery capabilities, executing an erroneous action will induce a cascading failure, trapping the system in an invalid reasoning loop until the maximum step limit is reached. Preemptive interception fundamentally prevents this severe waste of computational power and time.
    
    \item \textbf{WM Imagination:} Blocks invalid actions from entering the prediction network, preventing the high-latency WM from squandering valuable rendering compute budget and mitigating prediction deviations for future states.
\end{itemize}

By sharing the same verification gateway across physical execution and WM rollout, \method{} improves both safety and computational efficiency while preserving the generative capability of the base model. We next detail how \method{} is trained from safety-aware supervision signals (Section~\ref{expr:datasets}), implemented as a lightweight dual-branch network (Section~\ref{expr:train}), and deployed through a dual-mode runtime resampling scheduler (Section~\ref{expr:infer}).

\subsection{ Safety-aware Supervision Signal}
\label{expr:datasets}

In learning-based robotic arm control systems, relying directly on human annotations to define safety boundaries is not only prohibitively expensive but also inadequate for covering complex physical dynamics. Therefore, we propose extracting weak supervision signals from the online exploration rollouts of PPO~\cite{schulman2017proximal} Reinforcement Learning and reshaping them into strict system safety constraints.

\subsubsection{Chunk Advantage Estimation}
Specifically, we jointly model the language instruction $l$, visual observation $o_t$, proprioceptive state $q_t$, and the VLA-generated candidate action chunk $C_t=\{a_t^1, a_t^2, \dots, a_t^K\}$ as a multimodal context $x_t=\{l;o_t;q_t;C_t\}$. Based on this context, we employ a pre-trained PPO Critic network to compute the $K$-step guided chunk advantage:
\begin{equation}
\begin{aligned}
A_{t}^{K} &= \sum_{i=0}^{K-1}\gamma^{i}r_{t+i}
+ \gamma^{K} H_{t,K} V_{\psi}(x_{t+K})
- V_{\psi}(x_t); \\[1ex]
H_{t,K} &= \prod_{i=1}^{K} \big(1 - d_{t+i}^{\text{collision}}\big).
\end{aligned}
\end{equation}

Here, $\gamma \in (0,1)$ denotes the discount factor, whose value is kept consistent with that used during PPO training; $r_{t+i}$ is the single-step reward obtained from environmental feedback; and $V_{\psi}(\cdot)$ represents the PPO Critic value network. The term $H_{t,K}$ is a safety mask designed to truncate future value estimation when irreversible failures, such as collisions or object drops, occur during the execution of the action chunk. Specifically, if any $d_{t+i}^{\text{collision}}=1$, then $H_{t,K}=0$, thereby masking out the bootstrapped value term $V_{\psi}(x_{t+K})$ and preventing post-failure value signals from being propagated. In this way, the resulting supervision signal can better identify and penalize high-risk action chunks that may appear benign in the short term but lead to subsequent failures.

\subsubsection{Failure-Aware Advantage Refinement}

In embodied control scenarios, erroneous actions do not necessarily lead to immediate system failure. Instead, they may trigger catastrophic failures, such as collisions or object drops, only after several subsequent time steps. To model this causal delay and capture long-tail high-risk actions with delayed effects, we introduce a failure-aware penalty term $F_{t,K}$, which imposes an exponentially decayed penalty on the historical action window responsible for episode failure:
\begin{equation}
\begin{split}
    F_{t,K} &= \mathbb{I}[\text{episode fails}] \cdot 
    \mathbb{I}[0 \le T_{fail} - (t+K) \le W] \\
    &\quad \cdot \exp\left(-\frac{T_{fail} - (t+K)}{\kappa}\right).
\end{split}
\end{equation}

Here, $T_{fail}$ denotes the time step at which a catastrophic failure occurs, such as a collision or object drop; $W$ is the size of the failure backtracking window; and $\kappa$ controls the exponential decay rate of the penalty. Through this retrospective temporal mechanism, the penalty can be assigned to misleading actions that do not immediately trigger $H_{t,K}=0$ but may induce subsequent system failures. Accordingly, we revise the original advantage as:
\begin{equation}
\overline{A}_{t}^{K} = A_{t}^{K} - \lambda_{fail} F_{t,K}.
\end{equation}

In this way, the system-level penalty is incorporated into the original advantage, yielding the safety-aware advantage $\overline{A}_{t}^{K}$. Furthermore, in multi-task learning settings, different robotic manipulation tasks may exhibit varying reward baselines and value scales. Directly using the absolute advantage values would therefore lead to unstable monitoring thresholds. To address this issue, we standardize the safety-aware advantage within each task-level data buffer:
\begin{equation}
\tilde{A}_t^K = 
\frac{\overline{A}_t^K - \mu_{\text{task}}}
{\sigma_{\text{task}} + \epsilon}.
\end{equation}

Here, $\mu_{\text{task}}$ and $\sigma_{\text{task}}$ denote the mean and standard deviation of the advantage values in the current task buffer, respectively, and $\epsilon$ is a small constant used to prevent division by zero and ensure numerical stability.
\subsubsection{Threshold-Based Safety Label Generation}
Finally, we establish a unified safety pass threshold $\tau_A$ to convert the continuous safety evaluation signal into the binary supervision label $y_{t}^{binary}$ required by the \method{}, while simultaneously retaining the continuous fractional value as an auxiliary supervision label $y_{t}^{cont}$:
\begin{equation}
y_{t}^{binary} = \mathbb{I}(\tilde{A}_{t}^{K} \ge \tau_{A}), \quad y_{t}^{cont} = \tilde{A}_{t}^{K}
\end{equation}
where $\mathbb{I}(\cdot)$ is the indicator function. Through the rigorous signal refinement pipeline described above, we successfully transform the relative value signals originally intended for RL policy optimization into absolute safety classification boundaries suitable for runtime safety verification in \method{}.

\subsection{Lightweight Dual-Branch Runtime Monitor \method{}} %\yongjian{I am revising this}
\begin{figure}[t]
    \centering
    \includegraphics[width=\columnwidth]{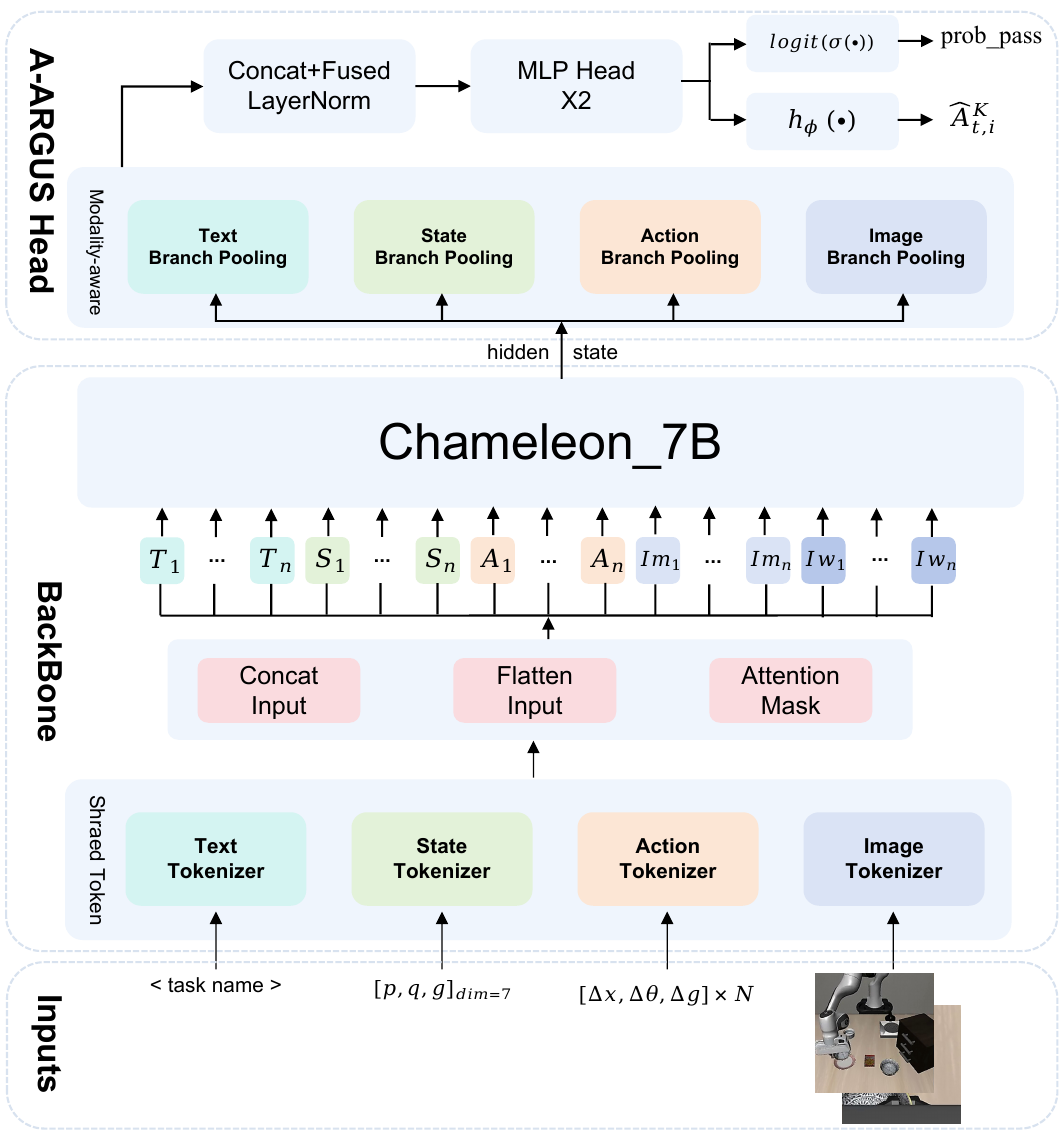}

    \caption{Overall architecture of \method{} with dual-modal data used during training.}
    \label{fig:monitor_architecture}
\end{figure}

\label{expr:train}
In robotic arm control systems, the monitor is required to perform verification with low computational overhead so as to support efficient online decision making. To this end, we design a lightweight monitor architecture, as shown in Fig.~\ref{fig:monitor_architecture}. The core idea is to reuse the multimodal perceptual representations already learned by the foundation model.

% \vspace{1em}
% \noindent \textbf{Multimodal Backbone and Feature Extraction:} 
\subsubsection{Multimodal Backbone and Feature Extraction}
Specifically, \method{} integrates an improved WorldVLA~\cite{cen2025worldvla} unified multimodal backbone. Built upon the vision-language model Chameleon~\cite{team2024chameleon}, this backbone introduces dedicated action and state tokenizers, thereby representing text instructions, visual observations, robot proprioceptive states, and candidate action chunks as a unified sequence of discrete tokens. In each planning cycle, the multimodal input sequence is fed into the backbone for encoding. During the training of \method{}, all parameters of the backbone are frozen to preserve its original generative capability. We then extract the final-layer hidden states $H_t$ of the backbone as high-dimensional feature representations for subsequent verification.

% \vspace{1em}
% \noindent \textbf{Modality-aware Pooling:} 
\subsubsection{Modality-Aware Feature Pooling}

Since the hidden states $H_t$ produced by the backbone integrate multimodal information from different input sources, modality-specific representations are highly entangled at the token level. To explicitly characterize the contribution of each modality to action validity assessment, we design a modality-aware pooling layer. Specifically, based on predefined modality masks $M_m$ ($m \in \{\text{text, image, state, action}\}$), this layer extracts the token representations corresponding to each modality from $H_t$ and applies Masked Mean Pooling independently to obtain modality-level summary vectors $z_m$:
\begin{equation}
    z_{m} = \frac{\sum_{j}M_{m}^{j}H_{t}^{j}}{\sum_{j}M_{m}^{j}+\epsilon}
\end{equation}

where $\epsilon$ is a numerical stabilizer used to avoid division by zero. The pooled representations of the four modalities are then concatenated to form a unified global feature representation $z_t = [z_{\text{text}}; z_{\text{image}}; z_{\text{state}}; z_{\text{action}}]$ for subsequent safety prediction. For missing modalities, the corresponding representation is filled with a zero vector to ensure consistent forward propagation.

% \vspace{1em}
% \noindent \textbf{Dual-branch Verification Head and Inference:}
\subsubsection{Dual-Branch Safety Prediction Head}
The concatenated global representation $z_t$ is fed into a minimalist dual-branch Feed-Forward Network (FFN). The primary classification branch is responsible for outputting the safety confidence $p_t \in [0, 1]$ of the candidate action chunk:
\begin{equation}
    p_t = \sigma(\text{FFN}_{\text{cls}}(z_t))
\end{equation}

where $\text{FFN}_{\text{cls}}(\cdot)$ denotes a two-layer lightweight FFN, and $\sigma(\cdot)$ is the Sigmoid activation function. Meanwhile, we introduce a parallel regression branch to predict the continuous chunk advantage derived from the Critic:
\begin{equation}
    \widehat{A}_t^K = h_\phi(z_t)
\end{equation}

where $h_\phi(\cdot)$ denotes an equally lightweight regression network. By directly reusing the backbone features and adopting a compact dual-branch FFN structure, this module requires only 100--200 ms of pure forward inference time in physical deployment. As a result, the system can perform proactive safety assessment and risk interception with low computational overhead before the future $K$-step action chunk is dispatched to the low-level controller for execution or fed into the WM for expensive rendering prediction.

% \vspace{1em}
% \noindent \textbf{Joint Training Objective:}
\subsubsection{Multi-Task Training Objective}
During the data collection process described in Section~\ref{expr:datasets}, safe action samples dominate the dataset, whereas high-risk actions that lead to task failures are extremely rare. Such severe class imbalance causes the standard cross-entropy loss to be biased toward a large number of easily classified safe samples, thereby weakening the model's ability to learn catastrophic failure cases. To address this issue, we formulate a multi-task joint loss function consisting of three sub-objectives:

\begin{itemize}
    \item \textbf{Focal Loss ($\mathcal{L}_{\text{cls}}$):} To address the severe class imbalance between safe and failure-inducing actions, we replace standard cross-entropy with Focal Loss:
    \begin{equation}
    \begin{split}
    \mathcal{L}_{\text{cls}} = {} & - y_{t}^{\text{binary}} \alpha_{\text{focal}} (1 - p_t)^{\beta_{\text{focal}}} \log(p_t) \\
    & - (1 - y_{t}^{\text{binary}}) (1 - \alpha_{\text{focal}}) p_t^{\beta_{\text{focal}}} \log(1 - p_t)
    \end{split}
    \end{equation}
    Here, $\alpha_{\text{focal}} \in [0,1]$ controls class balancing, and $\beta_{\text{focal}} \ge 0$ is the focusing parameter. The modulating factors down-weight easy safe samples and encourage the model to focus on rare hard negatives.
    
    \item \textbf{Regression Loss ($\mathcal{L}_{\text{reg}}$):} Combined with the regression branch, we formulate a Mean Squared Error loss to fit the normalized long-horizon system advantages:
    \begin{equation}
    \mathcal{L}_{\text{reg}} = \frac{1}{2} \left\| \widehat{A}_t^K - \tilde{A}_t^K \right\|_2^2
    \end{equation}
    Here, $\widehat{A}_t^K$ is the scalar predicted by the regression head, and $\tilde{A}_t^K$ is the normalized advantage target. This term regularizes the hidden representation $z_t$ with continuous dynamics-aware supervision.

    \item \textbf{Soft-threshold Loss ($\mathcal{L}_{\text{soft}}$):} Since the hard label $y_t^{\text{binary}}$ changes abruptly around the safety threshold $\tau_A$, we further construct a soft target from $\tilde{A}_t^K$ to smooth the decision boundary:
    \begin{equation}
        \begin{split}
            \mathcal{L}_{\text{soft}} = {} & - s_t \log(p_t) - (1 - s_t) \log(1 - p_t), \\
            & \text{with } s_t = \sigma\left(\frac{\tilde{A}_t^K - \tau_A}{\tau_{\text{temp}}}\right)
        \end{split}
    \end{equation}
    Here, $s_t \in (0,1)$ is the soft safety target, and $\tau_{\text{temp}} > 0$ controls the smoothness around the decision boundary. This loss improves confidence calibration for ambiguous near-threshold actions.

\end{itemize}

Finally, the joint optimization objective of \method{} is defined as:
\begin{equation}
    \mathcal{L} = \mathcal{L}_{\text{cls}} + \lambda_{\text{soft}} \mathcal{L}_{\text{soft}} + \lambda_{\text{reg}} \mathcal{L}_{\text{reg}}
\end{equation}
where $\lambda_{\text{soft}}$ and $\lambda_{\text{reg}}$ are balancing coefficients. This multi-task learning paradigm, which combines hard classification interception, continuous partial order regression, and soft boundary smoothing, ensures that \method{} can form an extremely stable and robust runtime safety decision boundary amidst highly non-linear and complex multimodal interactions.

\subsection{Dual-Mode Preemptive Resampling Scheduler}
\label{expr:infer}

After training \method{}, the system needs to seamlessly integrate \method{} into the inference pipeline of the foundation model at runtime. Considering the uncertainty of learning-based components, candidate actions may introduce potential safety risks or lead to ineffective resampling, thereby affecting execution reliability and computational efficiency. To address this issue, we design a dual-modal preemptive resampling scheduling mechanism to jointly support physical safety constraints and computational robustness during runtime.

To reduce the verification overhead in the early exploration stage and avoid prematurely constraining necessary preparatory actions, we introduce a system warm-up horizon $T_w$. When $t < T_w$, the system directly executes the candidate action; when $t \ge T_w$, \method{} is officially activated and incorporated into the verification pipeline:
$\text{use\_verifier} = \mathbb{I}[t \ge T_w]$.

Considering the fundamental difference between the underlying physics engine and the internal state prediction module, i.e., the Neural Simulator, in the execution system, we decouple the post-verification scheduling process into two independent modal subsystems: VLA execution in the physical actuator and World Model imagination.

\subsubsection{VLA Execution Scheduling in Physical Actuator}
In this modality, the candidate action chunks generated by the VLA are directly dispatched to the underlying physics engine, such as the LIBERO simulation environment, for dynamics execution. Meanwhile, to avoid repeated resampling caused by low-quality actions, we introduce a maximum retry number $N$ to constrain the scheduling process of \method{}.

\noindent \textbf{Preemptive Resampling:} The VLA first generates a candidate action chunk $a_{t,i}^K \sim \pi_\theta(\cdot \mid x_t; \xi_i)$ by injecting random noise $\xi_i$. The scheduler then invokes the classification branch of \method{} to compute its safety confidence $p_{t,i}$. If $p_{t,i} \ge \tau_p$, the action chunk passes verification and is dispatched to the physics engine for execution. If $p_{t,i} < \tau_p$, the action chunk is preemptively intercepted to avoid potential collisions or irreversible kinematic violations. When the resampling count satisfies $i < N$, the system instructs the VLA to inject new noise and generate another candidate action chunk.

\noindent \textbf{Regression-Guided Safe Fallback:} 
    When the number of resampling attempts reaches the upper limit $N$, the system terminates further resampling. If the classification confidences of all candidate actions are below the threshold $\tau_p$, the scheduler activates the regression branch and selects the action chunk with the highest continuous advantage score $\widehat{A}_{t,i}^K$ from the sampled candidates:
    \begin{equation}
        a_t^{K,*} = \arg\max_{i \le N} \widehat{A}_{t,i}^K
    \end{equation}
    The system then executes the selected candidate action chunk $a_t^{K,*}$ to maintain the continuity of the control process. This fallback mechanism prevents the system from falling into ineffective resampling loops while reducing the risk of dispatching low-quality actions to the physical execution process.

\subsubsection{World Model Imagination Scheduling}
The World Model is inherently a neural simulator. Its main challenges lie in the high GPU rendering cost and the cascading physical hallucinations induced by erroneous actions during forward propagation, namely misleading imagined futures. The exploration scheduling in this modality is likewise constrained by the maximum number of explorations $N$.

% \begin{itemize}
%     \item \textbf{Compute Protection and Interception:} Before the world model generates the future prediction $\hat{o}_{t+K} \sim f_\omega(\cdot \mid x_t, a_t^K)$, the candidate action must pass the classification score $p_{t,i}$. If $p_{t,i} < \tau_p$, the verifier prematurely rejects the action and refrains from generating future states for it. This mechanism fundamentally cuts off the squandering of invalid computational power, reserving computational resources for more valuable exploration paths.
    
%     \item \textbf{On-demand Truncation and Imagination Release:} If an action is intercepted, the system instructs the VLA to resample. When resampling reaches the upper limit $N$, the scheduler similarly utilizes the continuous regression branch $\widehat{A}_{t,i}^K$ to select the optimal candidate action to input into the WM. Specifically, under extremely adverse OOD states, if the optimal action still holds no long-term value, the system directly truncates the current Imagination Rollout, releasing GPU compute for global inference in the next control cycle.
% \end{itemize}

    \noindent  \textbf{Preemptive Compute Filtering:} Before the world model generates the future prediction 
$\hat{o}_{t+K} \sim f_\omega(\cdot \mid x_t, a_t^K)$, 
the candidate action is first evaluated by the classification branch of \method{}. If its safety confidence satisfies $p_{t,i} < \tau_p$, the action is rejected before being fed into the WM, and no future states are generated for it. This early interception avoids unnecessary world-model inference for unsafe candidates and reduces redundant GPU computation.
    
    \noindent \textbf{On-Demand Imagination Truncation:} If the candidate action is intercepted, the system guides the VLA to resample a new action chunk. When the number of resampling attempts reaches the upper limit $N$, the scheduler uses the regression branch to select the candidate action with the highest continuous advantage score $\widehat{A}_{t,i}^K$ from the sampled actions and feeds it into the WM. If, under complex OOD states, the selected candidate is still evaluated as highly risky, the system truncates the current imagination rollout to avoid further unreliable prediction and releases computational resources for the next control cycle.

\section{Experiment}

\subsection{Experiment Setup}

\noindent \textbf{Baseline Model:} We adopt RynnVLA-002~\cite{cen2025rynnvla} as the target VLA/WM baseline model. RynnVLA-002 contains both a VLA action generation branch and a world model future state prediction branch, making it suitable for evaluating \method{} in both closed-loop execution and WM imagination scenarios. \method{} is built upon an improved WorldVLA~\cite{cen2025worldvla} unified multimodal backbone and equipped with a retrained action verification head.

\noindent \textbf{Datasets:} The training data for \method{} is collected from PPO rollout~\cite{yu2025rlinfflexibleefficientlargescale} trajectories in LIBERO~\cite{liu2023libero}, where the PPO critic is used to construct action validity labels. The complete dataset contains 1,284,485 action chunk samples, with an initial positive-to-negative sample ratio of approximately 95:5, and is split into training and testing sets with a ratio of 92:8. Binary classification labels are generated from critic-derived scores, with the classification threshold set to $-0.21$. This threshold is also used for minority-class sample partitioning and soft-threshold loss construction. To alleviate the class imbalance problem, each training batch dynamically maintains a fixed 30\% proportion of negative samples. Other hyperparameters used during training and inference are summarized in Table~\ref{tab:hyperparameters}.

\begin{table}[htbp]
    \centering
    \caption{Hyperparameter configurations for \method{} training and inference.}
    \label{tab:hyperparameters}
    \begin{tabular*}{\columnwidth}{@{\extracolsep{\fill}}ll@{\hspace{1.0em}}|@{\hspace{1.0em}}ll@{}}
        \toprule
        \textbf{Parameter} & \textbf{Value} &
        \textbf{Parameter} & \textbf{Value} \\
        \midrule
        $K$                     & 5                  & $\gamma$                  & 0.99 \\
        $W$                     & 20                 & $\kappa$                & 3 \\
        $\lambda_{fail}$        & 1                  & $\epsilon$              & $1\times10^{-4}$ \\
        $\alpha_{\text{focal}}$ & 0.25               & $\gamma_{\text{focal}}$ & 2 \\
        $\lambda_{\text{reg}}$  & 0.05               & $\lambda_{\text{soft}}$ & 0.2 \\
        $\tau_{\text{temp}}$    & 0.25               & $T_w$                   & 20 \\
        $\tau_p$                & 0.275              & $N$                     & 5 \\
        lr                      & $3\times10^{-4}$   & Weight decay            & 0.01 \\
        Dropout                 & 0.1                & Epochs                  & 10 \\
        \bottomrule
    \end{tabular*}
\end{table}

\noindent \textbf{GPU Configuration:} The training of \method{} is conducted on a single server with 8$\times$ NVIDIA H20D GPUs, each with approximately 141 GB of memory. Training is launched in single-node DDP mode. The per-GPU training batch size is 512, and the model is trained for 10 epochs. The complete training process, including intermediate evaluations, takes approximately 8 days, 16 hours, and 43 minutes. The inference experiments are conducted on 16$\times$ NVIDIA RTX 5090 GPUs, each with 32 GB of memory. During inference, the VLA, \method{}, and WM are deployed on separate GPUs for subsequent experimental evaluation.

\noindent \textbf{Evaluation Metrics:} 
\begin{itemize}
    \item For closed-loop execution experiments, we report the success rate, average task steps, per-step inference time, and average attempts per step, which measure task completion ability, interaction efficiency, computational overhead, and the resampling cost introduced by \method{}, respectively.

    \item For offline \method{} evaluation, we report the F1 score, accuracy, invalid precision, invalid recall, false pass rate, and false reject rate to evaluate the model's ability to identify invalid actions, as well as its false-positive and false-negative tendencies.
\end{itemize} 

To ensure a fair comparison, all methods are evaluated under the same LIBERO simulation environment, RynnVLA-002 checkpoint, image preprocessing pipeline, action chunk length, maximum episode steps, random seed protocol, and candidate action sampling budget.

Our experimental evaluation is structured from multiple complementary perspectives. First, the action validity discrimination capability of the \method{} is validated on both the independent test set and successful/failed rollouts (Section~\ref{expr:discri}). Subsequently, the quantitative impact of the \method{} on VLA success rates and task steps is assessed via LIBERO suite-level closed-loop execution (Section~\ref{expr:success}). Following this, the visual role of the \method{} within world model imagination scenarios is analyzed via VLA-WM rollout visualizations (Section~\ref{expr:wm}). Finally, comprehensive ablation studies (Section~\ref{expr:abalation}) and specific case studies (Section \ref{expr:case}) are presented to isolate the individual contributions of each component and analyze performance variations across diverse specific tasks.

%\subsection{true fase}
\subsection{Discrimination Capability of the \method{} }
\label{expr:discri}

Before conducting the closed-loop execution evaluation, we first evaluate whether \method{} possesses reliable action validity discrimination capability on an independent test set. The detailed configuration of this test set is provided in Section~\ref{expr:success}. As shown in Table~\ref{tab:ablation_verification}, our method achieves an F1 score of 0.8303 and an accuracy of 0.9542 on the test set, indicating that \method{} can reliably distinguish between valid and invalid actions. In particular, \method{} achieves a precision of 0.7200 and a recall of 0.9800 on the invalid-action class, with a false pass rate of only 0.0200 and a false reject rate of 0.0491. These results show that \method{} can effectively intercept invalid actions while keeping the rejection rate of valid actions at a relatively low level.

Furthermore, we collect complete rollout trajectories of the VLA interacting with the LIBERO simulation environment, including 60 failed trajectories and 60 successful trajectories. For each trajectory, we compute the proportion of action chunks that are judged as valid by \method{}, so as to evaluate the correlation between \method{}'s predictions and the final task outcome. As shown in Table~\ref{tab:trajectory_pass_rate}, successful trajectories achieve an average pass rate of 0.8424, which is significantly higher than that of failed trajectories, 0.5167, with a gap of approximately 32.6\%. This result indicates that the predictions of \method{} not only characterize action validity at the action-chunk level, but also reflect execution quality at the trajectory level.

\begin{table}[htbp]
    \centering
    \caption{Action-chunk pass rates predicted by \method{} on successful and failed trajectories.}
    \label{tab:trajectory_pass_rate}
    \begin{tabular}{lccc}
        \toprule
        \textbf{Traj Type} & \textbf{Mean Pass Rate $\uparrow$} & \textbf{Min Pass Rate} & \textbf{Max Pass Rate} \\
        \midrule
        Failed      & 0.5167 & 0.4333 & 0.6036 \\
        Successful  & 0.8424 & 0.7931 & 0.9048 \\
        \midrule
        \textbf{Gap} & \textbf{+0.3257} & \textbf{+0.3598} & \textbf{+0.3048} \\
        \bottomrule
    \end{tabular}
\end{table}

Overall, these results demonstrate that the \method{} trained in this work can serve as a reliable action quality filtering module before subsequent VLA execution and world-model imagination, providing a dependable basis for intercepting low-quality actions and resampling candidate actions.

\subsection{Closed-Loop Execution Evaluation of \method{}}
\label{expr:success}

\begin{figure*}
    \centering
    \includegraphics[width=\textwidth, trim=7.1cm 5.3cm 10cm 4cm, clip]{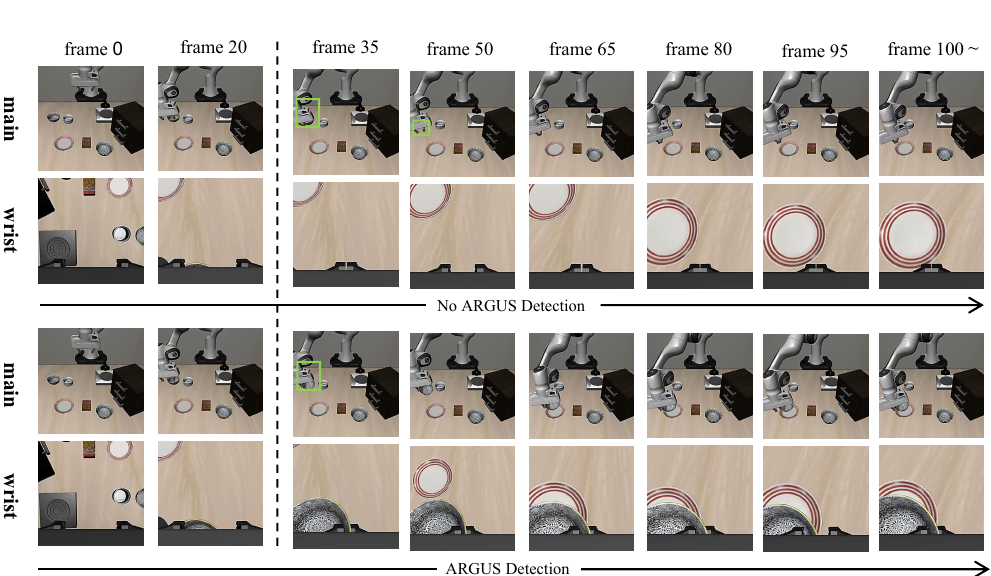}
    \caption{Closed-loop execution comparison with and without \method{}.}
    \label{fig:VLA-closed}
\end{figure*}

% \subsection{Closed-Loop Execution Evaluation of \method{}}
\label{expr:success}

\begin{figure*}
    \centering
    \includegraphics[width=1\linewidth]{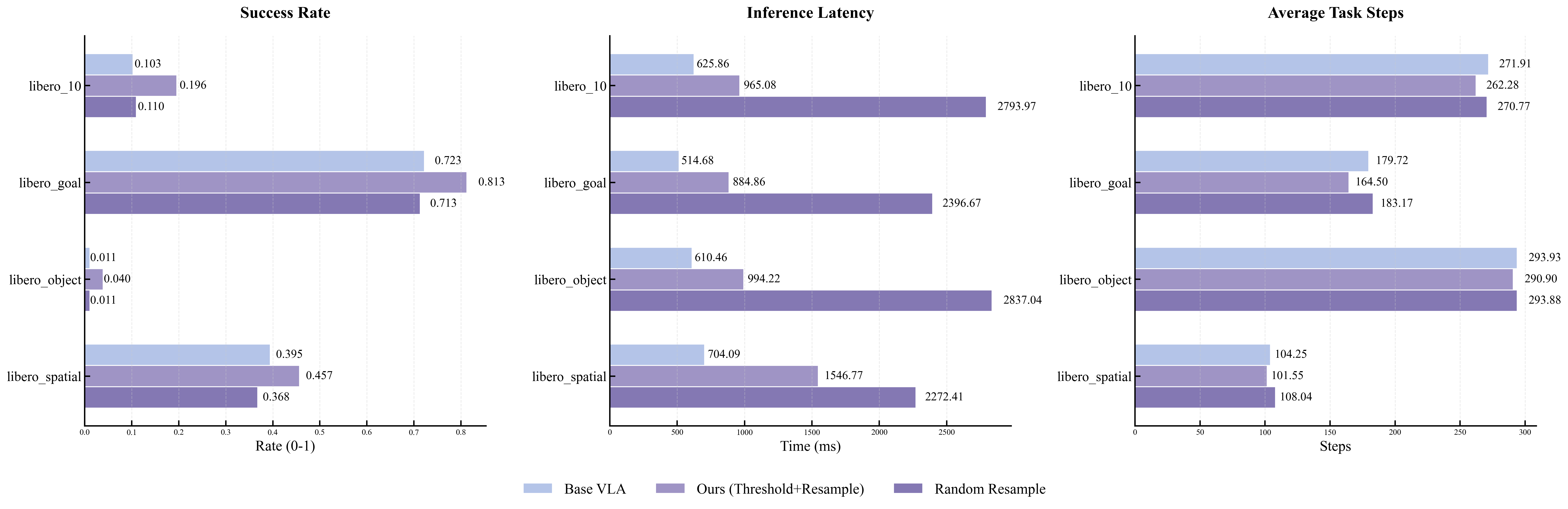}
    \caption{Closed-loop performance comparison across four LIBERO suites.}
    \label{fig:success}
\end{figure*}

\subsubsection{Suite-Level Performance Comparison}
To validate the role of \method{} in closed-loop execution, we integrate it into the VLA inference pipeline of RynnVLA-002 and evaluate it on four LIBERO suites, including LIBERO-10, LIBERO-Goal, LIBERO-Object, and LIBERO-Spatial. For each suite, we execute 10,000 episodes. We compare three methods: (1) RynnVLA-002, which serves as the baseline VLA and directly executes the first action chunk generated by the model; (2) Random Resampling, which randomly selects one action from the candidate action chunks for execution, in order to examine whether the performance improvement simply comes from additional sampling; and (3) \method{}, which evaluates the validity of candidate action chunks before execution. If a candidate action fails the verification, resampling is triggered until a valid action is obtained. The complete results are summarized in Figure~\ref{fig:success} and Table~\ref{tab:verifier_time}.

\noindent \textbf{Success Rate Analysis: }In terms of success rate (Figure~\ref{fig:success}), our method achieves the best performance across all four LIBERO suites. Compared with RynnVLA-002, \method{} improves the success rates on LIBERO-10, LIBERO-Goal, LIBERO-Object, and LIBERO-Spatial from 0.1029, 0.7225, 0.0114, and 0.3948 to 0.1961, 0.8126, 0.0395, and 0.4566, respectively. In contrast, the success rates of Random Resampling remain close to those of RynnVLA-002 and do not show consistent improvements. This indicates that the performance gain does not come from additional sampling alone, but from the effective discrimination of candidate action quality by \method{}. Notably, even in challenging scenarios such as LIBERO-Object, where the baseline success rate is relatively low, \method{} can still improve task success by filtering low-quality actions and resampling candidate actions to partially correct execution deviations.

\noindent \textbf{Task Step Efficiency: } Regarding the average number of task steps (Figure~\ref{fig:success}), our method improves the success rate without introducing longer execution trajectories. Specifically, the average numbers of task steps of \method{} on LIBERO-10, LIBERO-Goal, LIBERO-Object, and LIBERO-Spatial are 262.28, 164.50, 290.90, and 101.55, respectively, all lower than the corresponding results of RynnVLA-002. This suggests that \method{} improves performance not by extending the exploration process, but by filtering low-quality actions and helping the policy enter valid execution trajectories more efficiently.

\noindent \textbf{Resampling Attempts Analysis: } As shown in Table~\ref{tab:verifier_time}, the average numbers of attempts per step of \method{} on LIBERO-10, LIBERO-Goal, LIBERO-Object, and LIBERO-Spatial are 1.2841, 1.4216, 1.3696, and 2.0038, respectively. These results show that \method{} does not rely on excessive resampling to improve performance. In most cases, the model only needs one to two candidate action generations to obtain a valid action, indicating that \method{} can effectively select high-quality actions within a limited resampling budget.

\noindent \textbf{Closed-Loop Behavior Analysis: } Figure~\ref{fig:VLA-closed} illustrates the influence of \method{} on closed-loop execution. Without \method{}, RynnVLA-002 is more likely to generate target-deviating actions during the grasping stage, causing subsequent states to gradually move away from the task completion trajectory. In contrast, by filtering low-quality candidate actions before execution, our method reduces the probability of erroneous actions entering the environment. As a result, the robot arm exhibits more stable execution trajectories and moves more consistently toward the task goal.

\subsubsection{Runtime Efficiency Analysis}
Figure~\ref{fig:success} reports the per-step inference time of different methods. Since our method introduces \method{} scoring on top of VLA action generation and triggers resampling when candidate actions fail verification, its average per-step inference time is 1097.73 ms, higher than 613.77 ms of RynnVLA-002. Nevertheless, except for LIBERO-Spatial, where more complex spatial reasoning and a larger number of resampling attempts lead to inference time above 1000 ms, the per-step inference time on the remaining suites is controlled within 1000 ms. This generally satisfies the real-time requirements of closed-loop robotic manipulation. In addition, although Random Resampling samples five candidate actions at each decision step, its inference time does not increase linearly, suggesting that the resampling process reuses part of the existing context encodings or inference states.
\begin{table}[htbp]
    \centering
    \caption{Average forward time and resampling attempts of \method{} across LIBERO suites.}
    \label{tab:verifier_time}
    \begin{tabular}{lcc}
        \toprule
        \textbf{Suite} & \textbf{Average Forward Time (ms)} & \textbf{Average Attempts} \\
        \midrule
        LIBERO-Goal    & 178.1 & 1.4216 \\
        LIBERO-Spatial & 187.6 & 2.0038 \\
        LIBERO-Object  & 181.9 & 1.3696 \\
        LIBERO-10      & 188.0 & 1.2841 \\
        \bottomrule
    \end{tabular}
\end{table}

We further calculate the average time required for single-step action validity verification across different suites. As shown in Table~\ref{tab:verifier_time}, the average forward time of \method{} for verifying a single candidate action chunk is approximately 183.9 ms, indicating that the computational overhead of the verifier itself is relatively manageable. Therefore, the additional time cost of our method mainly comes from a small number of extra VLA sampling attempts rather than \method{} scoring. Overall, although \method{} introduces certain inference overhead, it leads to higher closed-loop success rates and fewer task execution steps, while still satisfying the real-time requirements of most tasks.

\subsection{World Model Rollout Analysis}
\label{expr:wm}
In the closed-loop VLA experiments above, action chunks are directly fed into the LIBERO simulation environment, and execution performance is evaluated through environment feedback. This section further verifies the applicability of \method{} in the internal VLA-WM interaction pipeline of RynnVLA-002. Specifically, the VLA branch first generates candidate action chunks, and the WM branch then performs rollout conditioned on these actions to predict a sequence of future state images. We insert \method{} between the VLA and the WM to evaluate the validity of candidate action chunks before they enter the WM for predictive rendering.

\begin{figure*}
    \centering
    \includegraphics[width=\textwidth, trim=2.7cm 1.5cm 2cm 1.5cm, clip]{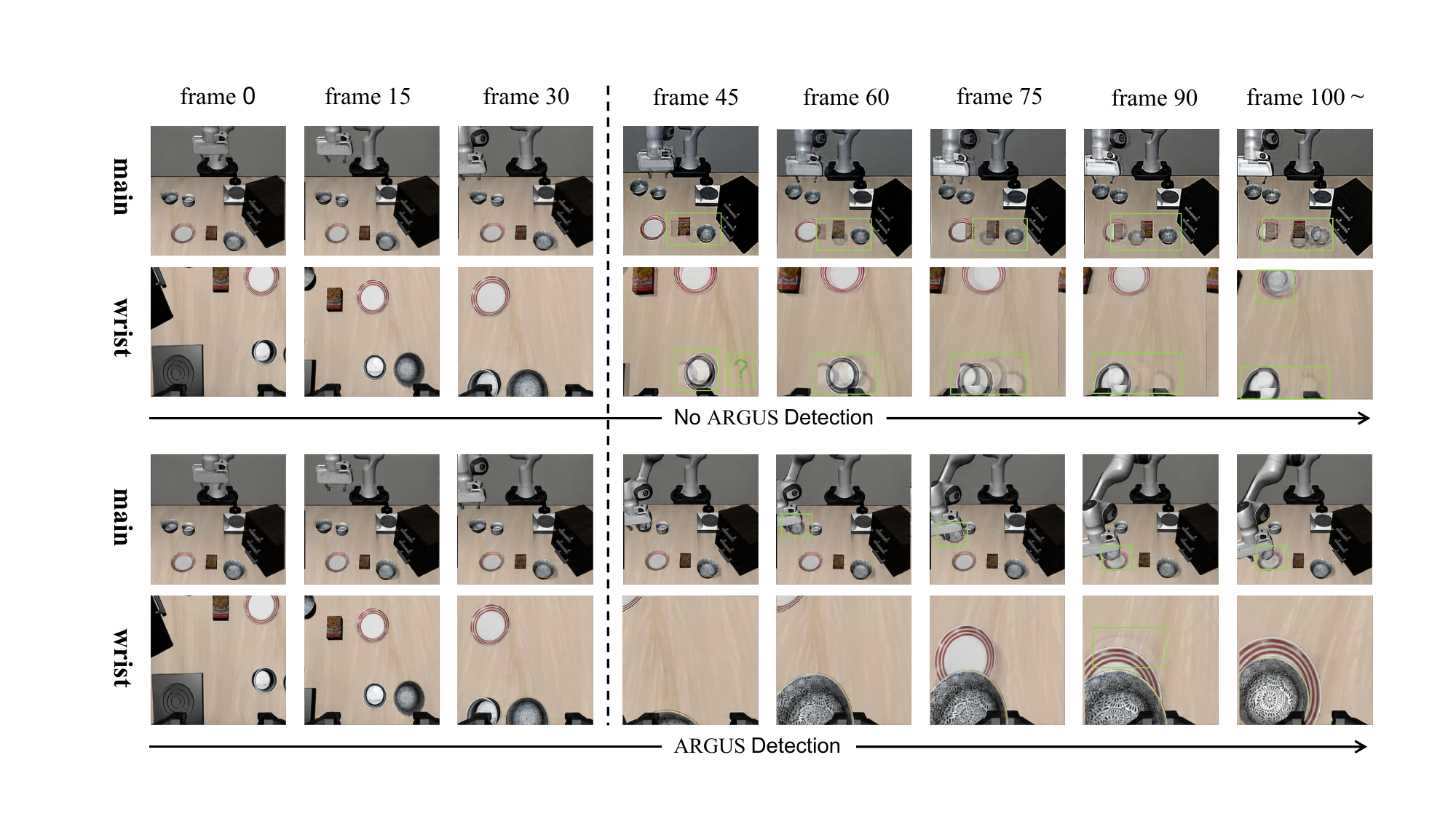}
    \caption{World Model rollout comparison with and without \method{}.}
    \label{fig:vla_wm_wide}
\end{figure*}

\begin{table*}[htbp]
    \centering
    \caption{Ablation study of action validity decision strategies.}
    \label{tab:ablation_verification}
    \begin{tabular}{lcccccc}
        \toprule
        \textbf{Method} & \textbf{F1} & \textbf{Accuracy} & \textbf{Invalid Precision} & \textbf{Invalid Recall} & \textbf{False Pass Rate} $\downarrow$ & \textbf{False Reject Rate} $\downarrow$ \\
        \midrule
        Raw Score Thresholding       & 0.4112 & 0.7006 & 0.2651 & 0.9163 & 0.0837 & 0.3271 \\
        Imbalance-aware Loss Scoring & 0.5078 & 0.7900 & 0.3466 & 0.9500 & 0.0500 & 0.2306 \\
        \textbf{Ours}                & \textbf{0.8303} & \textbf{0.9542} & \textbf{0.7200} & \textbf{0.9800} & \textbf{0.0200} & \textbf{0.0491} \\
        \bottomrule
    \end{tabular}
\end{table*}

Based on this setup, we select a representative pick-and-place task for visual analysis to compare the impact of introducing \method{} on WM rollout results. As shown in Figure~\ref{fig:vla_wm_wide}, the predictions under the two settings are largely consistent within the first 30 frames before the dashed line, where the motion trend of the robotic arm, the position of the target object, and the overall scene layout show no obvious deviations. However, after frame 30, the WM rollout without \method{} gradually exhibits noticeable errors: erroneous actions are directly passed to the WM, causing prediction errors to accumulate over time. This leads to target-deviating motions of the robotic arm, along with object position drift, blurring, and ghosting artifacts. As the rollout continues, these errors are further amplified and eventually result in a false completion that is inconsistent with real physical dynamics.

In contrast, with \method{} introduced, the system can filter low-quality actions before they enter the WM, thereby effectively suppressing error accumulation and propagation in the VLA-WM interaction loop. Although minor visual artifacts still appear in a few predicted frames, the overall scene structure and the spatial relationships among key objects remain stable. The robotic arm eventually follows a reasonable trajectory to complete the task, demonstrating that \method{} improves the stability and reliability of WM rollouts.

\subsection{Abalation Study}
\label{expr:abalation}

To analyze the impact of different training and decision strategies on the action validity \method{}, we conduct an ablation study on an independent test set. The test set contains 100,459 action chunk samples, with a positive-to-negative sample ratio of 8:2. We compare the following three settings:

\begin{itemize}
    \item \textbf{Raw Score Thresholding:} Directly uses the model output scores for threshold-based decision making.
    \item \textbf{Imbalance-aware Loss Scoring:} Introduces a loss function during training to alleviate the class imbalance problem, while still directly using the model output scores for threshold-based decision making during inference.
    \item \textbf{\method{}(Ours):} Built upon the imbalance-aware loss function, this method further maps the model output to an action pass probability and makes the final valid/invalid decision based on a probability threshold calibrated on the validation set.
\end{itemize}

As shown in Table~\ref{tab:ablation_verification}, directly applying thresholding to the raw scores achieves a relatively high invalid-action recall of 0.9163. However, its invalid-action precision is only 0.2651, and the false reject rate reaches 0.3271, indicating that many valid actions are incorrectly rejected. This is unfavorable for the stability of closed-loop execution. After introducing the imbalance-aware loss, the model performance improves: the F1 score increases from 0.4112 to 0.5078, the accuracy improves from 0.7006 to 0.7900, and the false pass rate decreases from 0.0837 to 0.0500. Nevertheless, since this setting still relies on uncalibrated model scores for decision making, its invalid-action precision remains relatively low at 0.3466, and the false reject rate is still 0.2306.

In contrast, our method combines the imbalance-aware loss with probability threshold calibration and achieves better performance across all metrics. Specifically, our method improves the F1 score to 0.8303 and the accuracy to 0.9542, while reducing the false pass rate and false reject rate to 0.0200 and 0.0491, respectively. These results indicate that the imbalance-aware loss helps mitigate the insufficient learning of invalid-action samples, while probability threshold calibration further aligns the model outputs with the final accept/reject decisions. Their combination more effectively reduces both the false pass of invalid actions and the false rejection of valid actions, thereby providing a more reliable action filtering signal for closed-loop VLA execution and world-model imagination.

\subsection{Case Study}
\label{expr:case}

In the aforementioned experiments (Section~\ref{expr:success}), we comprehensively evaluated the overall performance of the proposed method across different LIBERO suites. Since the current underlying VLA model is affected by prior cleaning of the original training data and thus achieves relatively low task success rates in suite-level evaluation, we further select six tasks on which RynnVLA-002 performs well as case studies to analyze the effectiveness of \method{} under a stronger baseline setting. For brevity, these selected tasks are denoted as Task 1 to Task 6, covering drawer opening, placing bowls or wine bottles at specified locations, and object pick-and-place operations in more complex scenes. Each task is evaluated over 250 episodes under two settings: RynnVLA-002 and Ours. The complete per-task results, together with the performance gains of Ours over RynnVLA-002, are summarized in Table~\ref{tab:task_gain}.

\begin{table}
    \centering
    \caption{Per-task success-rate gains and step reductions of \method{}.}
    \label{tab:task_gain}
    \small
    \begin{tabular*}{0.92\columnwidth}{@{\extracolsep{\fill}}ccc@{}}
        \toprule
        \textbf{Task ID} & \textbf{SR Gain (\%)} & \textbf{Avg. Step Reduction} \\
        \midrule
        1 & +6.86 & 2.7 \\
        2 & +4.43 & 3.6 \\
        3 & +2.04 & 1.2 \\
        4 & +2.81 & 0.6 \\
        5 & +5.96 & 3.3 \\
        6 & +3.15 & 8.1 \\
        \bottomrule
    \end{tabular*}
    \vspace{-0.8em}
\end{table}

Overall, our method \method{} achieves an average success rate of 90.72\% across the six tasks, outperforming RynnVLA-002, which achieves 86.51\%, by 4.21 percentage points. Meanwhile, our method requires an average of 171.70 task steps, compared with 174.83 steps for RynnVLA-002, reducing the average number of steps by 3.13. In addition, \method{} achieves an average decision time of 950.23 ms. These results indicate that even on tasks where RynnVLA-002 already achieves relatively high success rates, \method{} can further improve task success by filtering candidate actions, while reducing the average number of steps required to complete the tasks and satisfying the real-time requirements of online inference.

\section{Conclusion and Discussion}
\label{sec:conclusion}

In this paper, we introduced \method{}, a unified runtime verification architecture for improving the safety and efficiency of embodied robotic decision making. \method{} performs preemptive verification on candidate action chunks before they are dispatched to the physical environment for execution or used for world model imagination. Built upon an efficient unified multimodal backbone, \method{} combines modality-aware feature pooling with a lightweight dual-branch prediction head to jointly estimate action safety confidence and action advantage scores. In addition, we formulate a multi-task training objective that integrates Focal classification, advantage regression, and soft-threshold calibration to mitigate class imbalance and improve decision stability near safety boundaries.

Comprehensive evaluations on the LIBERO manipulation benchmark demonstrate the effectiveness of \method{} in action validity discrimination, closed-loop execution, world model rollout analysis, ablation studies, and task-level case studies. Specifically, compared with RynnVLA-002, \method{} improves the average closed-loop success rate across four LIBERO suites from 30.79\% to 37.62\%, while maintaining an average forward verification time of 183.9 ms per candidate action chunk. The results further show that \method{} reduces redundant execution steps and mitigates error accumulation in world model rollouts, suggesting that preemptive action verification can serve as an effective safety layer for foundation-model-based robotic control systems.

Despite these results, several limitations remain. First, the supervision signals used by \method{} are derived from PPO Critic estimates, which may introduce bias when the base policy encounters substantial distribution shifts. Second, although the forward verification overhead of \method{} is relatively low, repeated resampling under highly adverse out-of-distribution states may still accumulate additional inference cost. Future work will explore online self-calibrating verification mechanisms that adapt the safety boundary during interaction. We also plan to further investigate the generalization of \method{} across more diverse robot embodiments, task distributions, and open-world manipulation scenarios.

\bibliographystyle{IEEEtran}
\bibliography{mybib}

% \vspace{12pt}
% \color{red}
% IEEE conference templates contain guidance text for composing and formatting conference papers. Please ensure that all template text is removed from your conference paper prior to submission to the conference. Failure to remove the template text from your paper may result in your paper not being published.

\end{document}